\documentclass[10pt, conference, compsocconf]{IEEEtran}
\usepackage{cite}
\usepackage{upgreek}
\usepackage{mathtools}
\usepackage{gensymb}
\usepackage{tabu}
\usepackage[bf]{caption}
\usepackage{amsmath}
\usepackage{algorithm}
\usepackage[noend]{algpseudocode}
\usepackage{mathtools}
\usepackage{amsfonts}
\usepackage{makecell}
\usepackage{breqn}
\usepackage{subcaption}
\usepackage{caption}
\usepackage{float}
\usepackage{float}

\DeclareMathOperator*{\argmax}{arg\,max}
\DeclareMathOperator*{\argmin}{arg\,min}

\makeatletter
\def\BState{\State\hskip-\ALG@thistlm}
\makeatother

\begin{document}
\title{Shared Multi-Task Imitation Learning for Indoor Self-Navigation}
\author
{\IEEEauthorblockN{Junhong Xu, Qiwei Liu, Hanqing Guo, Aaron Kageza, Saeed AlQarni, Shaoen Wu}
\IEEEauthorblockA{Computer Science\\ Ball State University\\}
Muncie, IN USA \\
\{jxu7, qliu4, hguo, agkageza, saalqarni, swu\}@bsu.edu
}

\maketitle

\begin{abstract}
Deep imitation learning enables robots to learn from expert demonstrations to perform tasks such as lane following or obstacle avoidance. However, in the traditional imitation learning framework, one model only learns one task, and thus it lacks of the capability to support a robot to perform various different navigation tasks with one model in indoor environments. This paper proposes a new framework, \textit{Shared Multi-headed Imitation Learning}(\textit{SMIL}), that allows a robot to perform {\it multiple} tasks with {\it one} model without switching among different models. We model each task as a sub-policy and design a multi-headed policy to learn the shared information among related tasks by summing up activations from all sub-policies. Compared to single or non-shared multi-headed policies, this framework is able to leverage correlated information among tasks to increase performance. We have implemented this framework using a robot based on NVIDIA TX2 and performed extensive experiments in indoor environments with different baseline solutions. The results demonstrate that \textit{SMIL} has doubled the performance over non-shared multi-headed policy.
\end{abstract}

\section{Introduction} \label{intro}
One of the main challenges in robotics is to enable robots to interact with a dynamically changing environment and to perform different tasks with minimal prior knowledge. It requires the robot to perceive the environment, understand the context of the tasks, and make decisions accordingly. Traditional methods rely on accurate manual modeling for each task, such as visual SLAM systems for navigation \cite{mur2015orb}. In contrast, deep learning based methods simplify the need of manual modeling. They directly learn a policy that maps a sensor input to a corresponding control command. Because of the advance of deep learning models, especially convolutional neural networks (CNNs), learning control commands directly from environmental images becomes feasible \cite{Lecun2015deep}. 

However, literature imitation learning solutions can only learn one task per model. This prevents robots from executing complex actions. For example, if the robot is asked to fetch a cup in the kitchen, it needs to decompose the task into a few sub-tasks such as \textit{goto}, \textit{traverse}, and \textit{fetch}. In this work, we propose a framework that solves the two navigation problems with only one model, \textit{goto} and \textit{traverse}. The framework is called  \textit{\textbf{S}hared \textbf{M}ulti-task \textbf{I}mitation \textbf{L}earning (SMIL)},  which gives robots the ability to perform different navigation tasks in indoor environments. The framework is based on the multi-task learning (MTL) \cite{caruana1998multitask} and aims at solving task agnostic problem of imitation learning. Although MTL has been researched for a long time, multi-task imitation learning has rarely been researched until recently \cite{hausman2017multi, duan2017one}. 

The proposed framework uses a shared CNN to learn an environment model that extracts environmental features. Different sub-task policies are represented by a multi-headed fully connected network whose inputs are from the last layer of the shared CNN. While current literature solutions such as the work \cite{codevilla2017end} do not consider the relevance between the sub-tasks, our proposed framework rather makes use of the relevant information among sub-tasks. In order to solve poor generalization and distribution mismatch problems, we apply off-policy imitation learning \cite{laskey2017dart}, data augmentation, and dropout \cite{srivastava2014dropout} during training. During testing, the framework switches between sub-policies based on human navigation commands.

Our contributions are as below:
\begin{itemize}
\item We propose a new network architecture that leverages task relationships by summing up activations from sub-policies.
\item Off-policy learning procedure is used to train our framework.
\item Dropout and image augmentation are used to improve generalization.
\end{itemize}

In the rest of this paper, Section \ref{sec:related} reviews the related work of the imitation learning and multi-task learning. Our solution \textit{SMIL} framework is described in Section \ref{sec:smil} including the network architecture and detailed training procedure. Next, Section~\ref{ref:exp} presents extensive performance evaluations of {\it SMIL} in real indoor environments. The conclusion and future work are presented in Section~\ref{sec:con}.

\section{Related Work}\label{sec:related}
Learning based algorithms have been applied to a variety of robotics control problems. These algorithms can learn an end-to-end controller directly from data. For example, in \cite{Zhu2016Target}, reinforcement learning is used to train a siamese neural network to navigate to a target position. Similarly, the work \cite{mirowski2016learning} uses auxiliary losses to train a reinforcement learning agent to navigate through complex maze environments. In contrast to reinforcement learning, imitation learning has been applied to many real-world applications such as robotic grasping \cite{Laskey2016Robot}, UAV flight control \cite{giusti2016machine}, self-driving cars \cite{bojarski2016end}, and  rope manipulation \cite{nair2017combining}. 

The above mentioned algorithms only consider completing one task at a time, but this is not enough for many robotic tasks. Therefore, we target on the idea of multi-task learning (MTL) \cite{ruder2017overview}. Researchers have proposed a learning architecture that uses one single model to jointly learn image classification, speech recognition, and translation problems and yielded encouraging results\cite{kaiser2017one}. Long et. al. place a matrix prior to fully connected layers in a CNN to learn the relations of multiple tasks \cite{long2015learning}. Multi-task imitation learning has drawn attentions in recent years including learning multiple tasks together \cite{hausman2017multi, codevilla2017end} or one-shot learning \cite{santoro2016one, finn2017one, duan2017one}. Among these works, two works \cite{codevilla2017end, hausman2017multi} are most similar to ours. Hausman et. al. propose a multi-model imitation learning framework that separates video segments into different skill trajectories and imitate the demonstrated skills jointly\cite{hausman2017multi}. In \cite{codevilla2017end}, the authors propose a framework that learns sub-policies using a multi-headed network in the autonomous driving setting. However, these works have not considered the relevant information among tasks. Our proposed framework learns the relationships across tasks by combining learned features across sub-policies. By learning these relationships, the model is able to yield a more general representation of various navigation tasks.

\section{Shared Multi-task Imitation Learning}\label{sec:smil}
In this section, we first formally define the problem of imitation learning and multi-task imitation learning. Next, we present our deep learning network architecture. Finally, our training procedure is described in detail. 
\subsection{Problem Formulation}

To formally define the problem, let $\mathcal{X}$ and $\mathcal{Y}$ denote recorded observations and corresponding expert control commands respectively. A pair of ($\mathcal{X}^k$, $\mathcal{Y}^k$) is defined as the $k$-th {\it demonstration} of a robot. To simplify the notation, we assume the environment is Markovian; namely the current observation includes all history information of the environment. 

\subsubsection{Traditional Imitation Learning}
In the traditional imitation learning setting, the demonstrations are associated with a single task.  Thus, $\mathcal{X} = \{ \mathcal{X}^1, \mathcal{X}^2, ..., \mathcal{X}^n\}$ and $\mathcal{Y} = \{ \mathcal{Y}^1, \mathcal{Y}^2, ..., \mathcal{Y}^n\}$ represent $n$ demonstrations of the task. The imitation learning aims to learn a policy that maps an observation to a probability distribution of control command $\pi: \mathcal{X} \rightarrow \pi(\mathcal{Y}; \theta)$, where $\theta$ denotes the parameters of a weight vector, e.g. neural network. The parameters $\theta$ can be found by solving a maximum-likelihood estimation (MLE) problem: $\theta^* = \underset{\theta}{\argmax}\sum_{n=1}^{N} \pi({\mathcal{Y}^n|\mathcal{X}^n; \theta})$. If the policy $\pi$ is Gaussian and it is parameterized by a weight vector $\theta$, the MLE objective can be transformed into a $l2$ regression problem: $\theta^* = \underset{\theta}{\argmin}\frac{1}{N}\sum_{n=1}^{N}|| \pi({\mathcal{X}^n; \theta}) -\mathcal{Y}^n ||_2^2 $. In our system, we assume the policy is Gaussian thus we use the $l2$ objective to optimize the parameters $\theta$. 

\subsubsection{Multi-Task Imitation Learning}
In the multi-task imitation learning setting, observations $\mathcal{X}$ and the corresponding control commands $\mathcal{Y}$ are representing more than a single task in demonstrations. Instead, they consist of multiple demonstrated tasks and are denoted as $\mathcal{X} = \{X_1, X_2, ... X_i\}$ and $\mathcal{Y} = \{Y_1, Y_2, ..., Y_i\}$, where $X_i = \{X_i^1, X_i^2, ..., X_i^n\}$ and $Y_i = \{Y_i^1, Y_i^2, ..., Y_i^n\}$ represent the observations and control commands of task $i$. It should be noted that although here we use the same number of demonstrations across tasks to simplify the notations, it is not required to be the same. In addition, we introduce the concept of task embedding denoted as $T= \{T_1, T_2, ... T_i\}$. Similar to word embedding\cite{levy2014neural}, the task embedding is a feature vector for each task that embeds the high-level representations into a low-dimensional space.Therefore, the policy $\pi$ maps observations $\mathcal{X}$ and task embedding $T$ to a distribution of control command $\mathcal{Y}$ and is formulated as $\pi: \{\mathcal{X}, T\} \rightarrow \pi(\mathcal{Y}; \theta)$. Instead of minimizing the objective for one task, multi-task imitation learning aims to find a set of parameters that are to be optimized over multiple tasks: 
\begin{equation}\label{eq:mtil}
\theta^* = \underset{\theta}{\argmin}\frac{1}{I}\frac{1}{N}\sum_{i=1}^{I}\sum_{n=1}^{N}|| \pi({\mathcal{X}^n_i, T_i; \theta}) -\mathcal{Y}^n_i ||_2^2,
\end{equation}
where $I$ is the number of tasks. 

\subsection{Network Architecture}
With the multi-task imitation learning problem formulated as above, we have designed a  \textit{\textbf{S}hared \textbf{M}ulti-task \textbf{I}mitation \textbf{L}earning} (SMIL) framework that learns to perform four tasks based on human commands. This framework is shown in Fig. \ref{fig:multi_headed_shared}, which consists of two modules: image feature extractor and shared multi-headed policy, which will be explained in details in the following. 
\begin{figure*}[h]
\centering
\hspace{0.2in}
\includegraphics[width=7in]{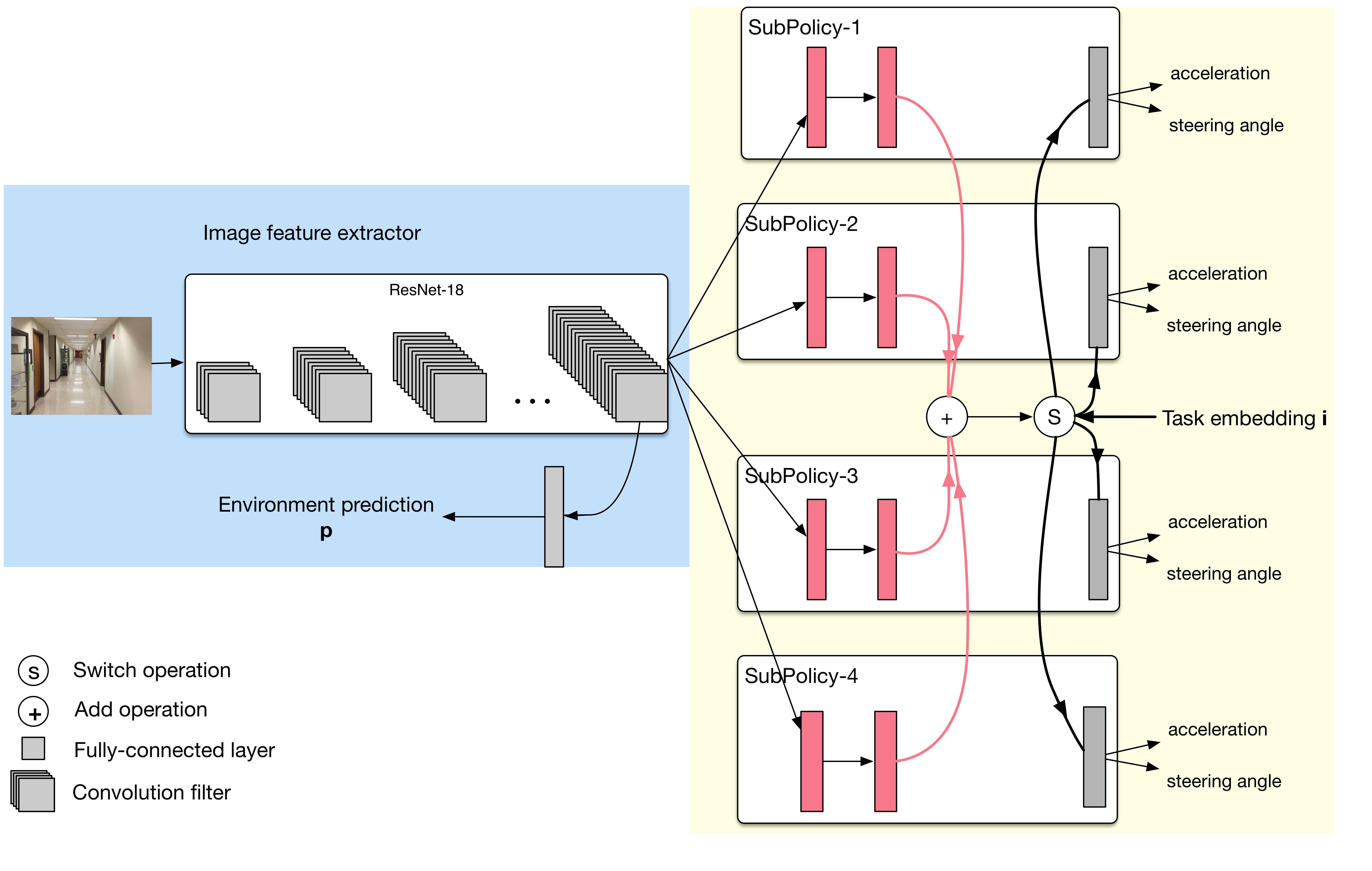}
\caption{An illustration of SMIL framework. An image observation is first passed through a ResNet-18 to extract features. The features are then passed through a linear classifier as well as a shared multi-headed policy to predict environment labels and control commands.}
\label{fig:multi_headed_shared}
\end{figure*}

\subsubsection{Image Feature Extractor}
The image feature extractor is a fine-tuned and pre-trained ResNet-18 by excluding the classifier layer in the original ResNet-18, but preserving the average pooling layer. It is used to project raw image inputs to a low-dimensional feature space and is denoted as $f = I(x, \theta_I)$, where $f\in \mathbb{R}^{1\times 1 \times 512}$ is the extracted feature vector, $x$ is the input image, and $\theta_I$ is the parameter set of ResNet-18. The idea of using pre-trained model has been extensively researched by the literature and it is proven to have faster convergence than training from scratch \cite{huh2016makes}. The feature vector $f$ is then flattened and passed to the shared multi-headed policy to generate control commands. In addition, to learn a more general representation of indoor environments,  image feature extractor also predicts which indoor environment the robot is currently in: $p = E(f)$, where $E$ represents a linear classifier. In our case, it predicts two class labels: hallway and classroom. Environment prediction is jointly trained with control commands.

\subsubsection{Shared Multi-headed Policy}
The shared multi-headed policy learns shared knowledge across tasks as well as task specific knowledge. As shown in Figure \label{fig:multi_headed_shared}, the shared multi-headed policy consists of three parts: switch operation, addition operation, and sub-policies represented by fully connected layers. In our case, we define four tasks to be learned: {\it traverse hallway, traverse classroom, to hallway,} and {\it to classroom}. We denote the shared multi-headed policy as $a = \pi(f, T; \theta_{\pi})$, where $\theta_\pi = \{\mathcal{W}^1, ..., \mathcal{W}^4\}$ is the parameter of the entire policy and $\mathcal{W}^i$ is the parameter set of the $i$th sub-policy. It takes two inputs, extracted features $f$ from the image feature extractor module and task embedding $T$, and gives the output of the control commands $a$ corresponding to a specific task. In our case, since we have four tasks, we denote the task embedding $T$ as a 4-dimensional one-hot vector\cite{codevilla2017end}. It is used to determine the sub-policy to be activated. Note that $T$ can also be learned in an unsupervised way \cite{hausman2017multi}. The action space is two-dimensional: acceleration and steering angle. 

Each sub-policy is a three-layer fully connected neural network. Because the tasks are highly correlated, it is useful to learn the task relationship through which the activated sub-policy can exploit useful information from other sub-policies. For example, the hallway navigation sub-policy can leverage obstacle avoidance knowledge learned by the classroom navigation sub-policy because classroom is a more complicated environment with different types of obstacles. Formally, denote $\mathcal{W}^i_l$ the set of parameters at layer $l$ in $i$-th sub-policy and $h^i_l = g_l(\mathcal{W}^{i^T}_lx^i_l)$ the corresponding output, where function $g_l(\cdot)$ denotes a non-linear function and $x^i_l$ is the input of that layer. Our network uses ReLU for the first two layers' non-linear functions and an identity function for the last layer. The input to first layers of the sub-policy networks is the extracted features $f$ from the image feature extractor module. The information across sub-policies is shared by using an addition operation that combines all the outputs from the second layers of each sub-policy network: 
\begin{equation}\label{eq:add}
j = \sum_{i=1}^{L}{h^i_2},
\end{equation}
 where $L$ is the total number of tasks. Although the literature has shown that higher layers learn more task specific features that are difficult to transfer \cite{yosinski2014transferable} , we choose to share the information from the second layers, because the tasks are highly correlated and the learned features are easier to transfer over sub-policies. After the addition operation, the task embedding $T$ selects a sub-policy to use via a switch operation. The switch operation routes the output from the addition operation to the final layer of the selected sub-policy. The final output will be:
 \begin{equation}
 a = h_3^t = g_3^t(W_3^t j).
 \end{equation}
 
 This design enables each sub-policy to learn the task specific controls in the final layer as well as to share knowledge through the addition operation across different sub-policies.

\subsection{Training Procedure}
% dropout, data augmentation, noise injection
It is important to train the multi-task imitation learning framework for high performance. We employ three different training techniques to train a robust SMIL framework: dropout, data augmentation, and noise injection.  Because \textit{SMIL} predicts steering angle and acceleration, which are real-valued numbers, we use mean squared error ($MSE$) as the loss function during the training.

\subsubsection{Dropout and Data Augmentation}
\begin{figure}[htp]
\begin{center}
\begin{tabular}{cc}
\includegraphics[scale=0.14]{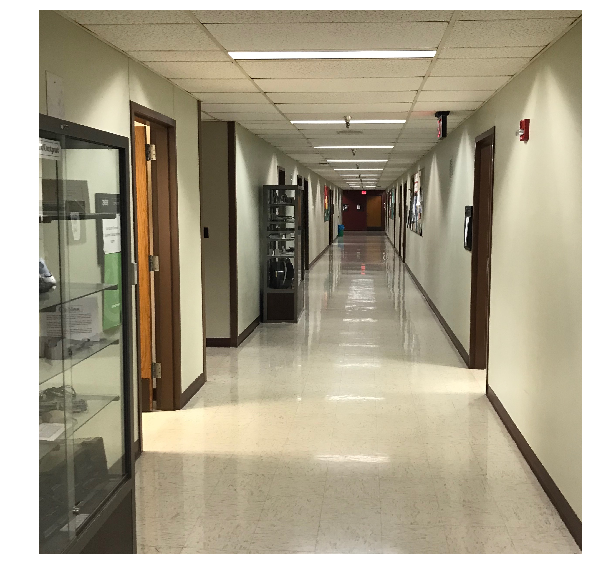}&
\includegraphics[scale=0.14]{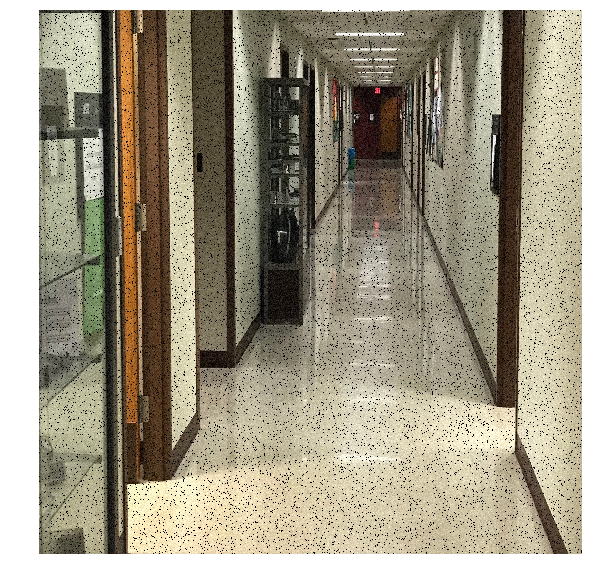}\\
\includegraphics[scale=0.14]{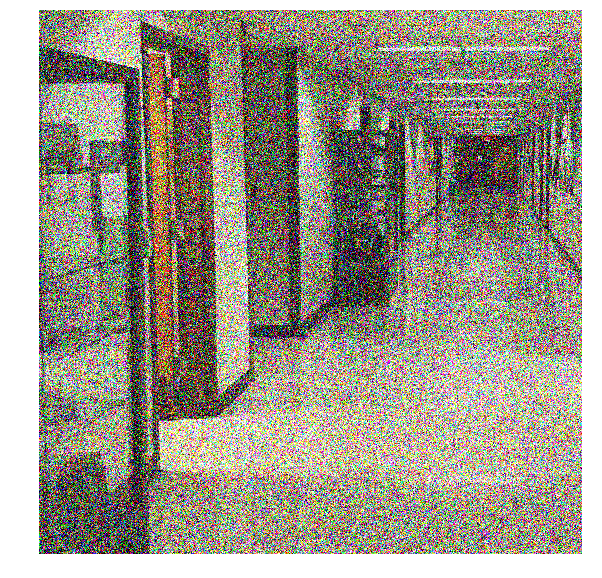}&
\includegraphics[scale=0.14]{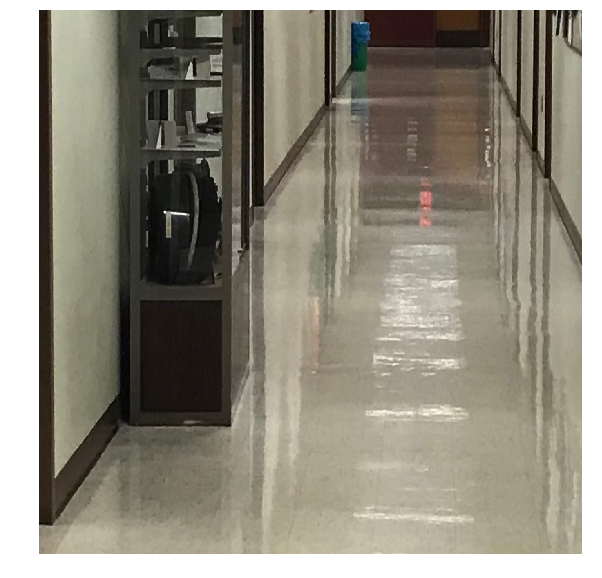}
\end{tabular}
\end{center}
\caption{An illustration of randomly chosen augmentation strategies. From left to right are original, random cropping and pixel dropout, additive Gaussian noise and contrast changing, and more aggressive random cropping.}
\label{fig:aug}
\end{figure}

As opposed to \cite{tobin2017domain} that is designed for single task frameworks, our goal is to train a robust SMIL framework that is able to perform in different environments from real-world experience, instead of learning from images generated by a hand-engineered simulator. It is necessary to collect images from diverse indoor environments to prevent a deep learning model from overfitting, but collecting data is time consuming. Hence, we use data augmentation and dropout \cite{srivastava2014dropout} to train a robust model. For data augmentation, we randomly apply contrast change, Gaussian noise, pixel dropout, random cropping, and horizontal flip (steering angle is also flipped). Different augmentation strategies are shown in Fig. \ref{fig:aug}. Dropout is used to prevent model overfitting by randomly zeroing out an neuron's activation. In addition, dropout can also stabilize the performance of the robot. Because of the addition operation, the norm of activation input to the final layer is possible to be very large and yields the outputs of very different control commands. Dropout is only added to the first and second layers of sub-policies. 

\subsubsection{Noise Injection}
Distribution mismatch between the supervised training and the robot learning is an essential problem in imitation learning: because the human operator is proficient in demonstrating tasks, there are no demonstrations of recovering from dangerous or erroneous states. As a result, the robot does not know how to correct itself in experiencing these abnormal states. We explore the off-policy training for the SMIL framework to address this issue, where the robot learns from another policy. We employ the noise injection approach \cite{laskey2017dart}, which injects an optimized Gaussian noise to an expert policy to maximize the probability of a human demonstrator making the same mistakes as the robot. 

Given a robot policy $\pi_{\theta}$, an expert policy $\pi^*$, DART algorithm \cite{laskey2017dart} aims to find a covariance matrix of Gaussian noise that maximizes the probability of expert taking the robot policy: 
\begin{align}
\begin{split}
\Sigma^* = \underset{\Sigma}{\argmin} \  E_{p(\upxi|\pi^*,\Sigma)} -\sum_{t=0}^{T-1}log[\pi*(\pi_{\theta}|x_t, \Sigma)]
\end{split}
\end{align}
where $\upxi$ means trajectories encountered by executing the expert policy with noise injected and $\Sigma$ is the covariance matrix. A shrinkage estimation is then utilized to scale the covariance matrix and derive a closed form solution:
\begin{align}
\begin{split}
&\Sigma^{\alpha} = \frac{\alpha}{Ttr(\Sigma^*)}\Sigma^*; \  \\
&\Sigma^*=\frac{1}{T}E_{p(\upxi|\pi^*,\Sigma)}\sum^{T-1}_{t=0}(\pi_{\theta}(x_t) - \pi^*(x_t))(\pi_{\theta}(x_t)-\pi^*(x_t))^T
\end{split}
\end{align}
where $\alpha$ is the prior knowledge of the final error of the robot policy on the training dataset. This algorithm is best used in an iterative approach, so we collect expert demonstrations for $k$ iterations and update the covariance matrix at the start of every iteration except for the first iteration with the covariance matrix as 0. In our case, we found $\alpha=2$ gives the best results, which is reasonable because we normalize value of steering angle and acceleration between -1 and 1, and the largest $MSE$ should not exceed 4.

\section{Performance Evaluation}\label{ref:exp}
We have implemented our framework and elevated in real world environments. Our experiments have been designed to answer the following questions: 
\begin{enumerate}
\item Is the shared task representation (the addition operation) necessary to the multi-task imitation learning when the tasks are highly correlated?
\item What is the performance difference of the multi-headed sub-policy framework compared to a single-headed policy?
\item Does the environment prediction task improve the performance?
\item Are data augmentation and dropout important to the model generalization and robustness?  
\end{enumerate}

\subsection{Testbed, Experiment Environment}
Our framework is implemented into an iRobot Create2 robot. The linear speed of this robot is in the range of -0.5 m/s to 0.5 m/s and the angular velocity is in the range of -4.5 rad/s to 4.5 rad/s, where a negative linear speed represents moving backward and a negative angular speed represents going right. The only sensory data we used are RGB images from a ZED Stereo camera. The valid depth estimation is between 0.5m and 20m. We use a NVIDIA Jetson TX2 as the main computation resource to do inference.

\begin{figure}[t]
% \hspace{-0.35in}
  \begin{tabular}[c]{cc}
    \begin{subfigure}[c]{0.23\textwidth}
    	\centering
     	\includegraphics[width=1.5in, height=1.0in]{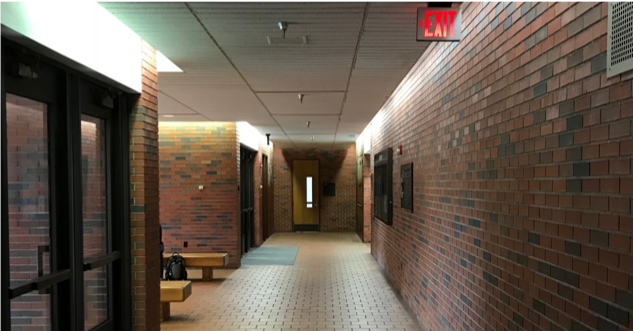}
     	\caption{Robert Bell first floor hallway.}
    \end{subfigure}&
    \begin{subfigure}[c]{0.23\textwidth}
    	\centering
	\includegraphics[width=1.5in, height=1.0in]{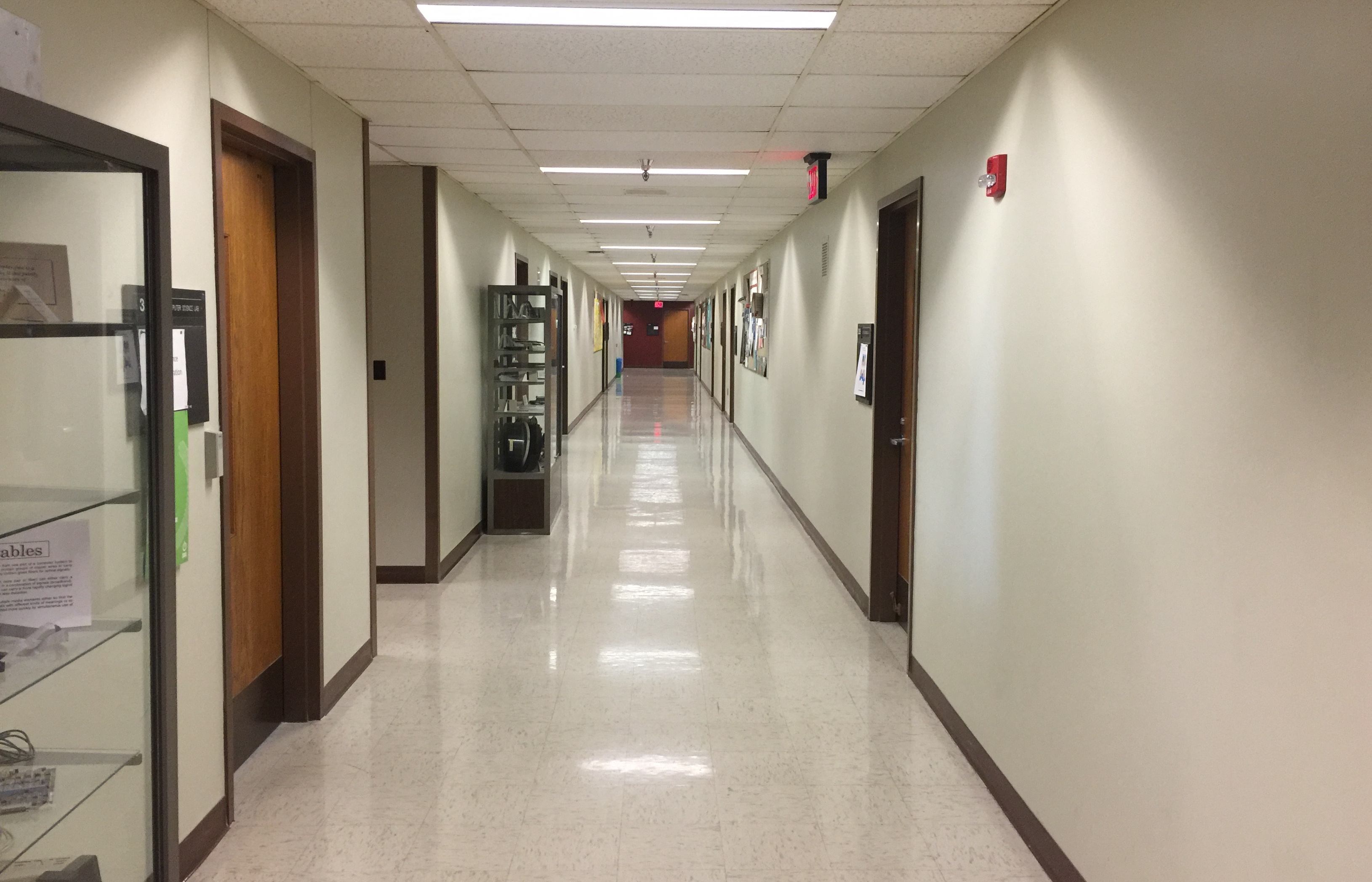}
	\caption{Robert Bell third floor hallway.}
    \end{subfigure}\\
        \begin{subfigure}[c]{0.23\textwidth}
    	\centering
     	\includegraphics[width=1.5in, height=1.0in]{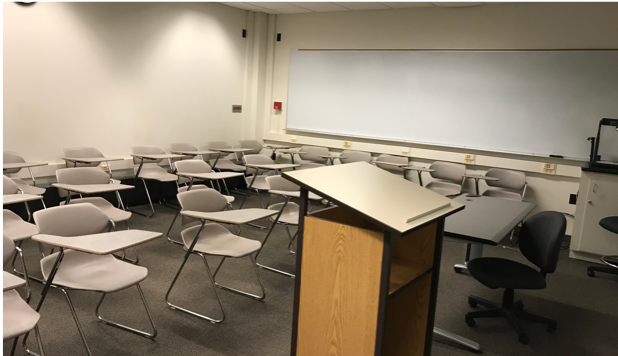}
     	\caption{Robert Bell first floor classroom.}
    \end{subfigure}&
    \begin{subfigure}[c]{0.23\textwidth}
    	\centering
	\includegraphics[width=1.5in, height=1.0in]{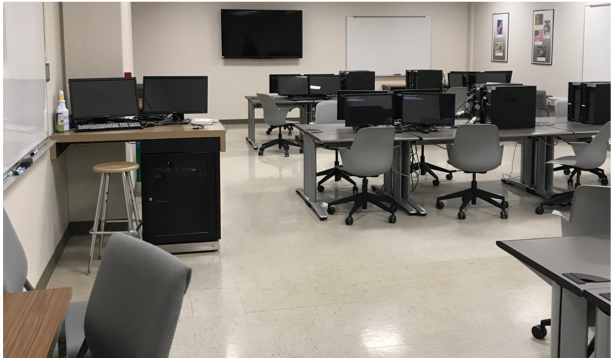}
	\caption{Robert Bell third floor classroom.}
    \end{subfigure}
  \end{tabular}    
\caption{Training environments (b) and (d) have different geometric and color appearances as testing environments (a) and (c).}
\label{fig:smil_exp}
\end{figure}

We have extensively evaluated our solution in a real environment: the Robert Bell Hall building at Ball State University. To test the generalization, we have trained the robot in the third floor but tested the robot on the first floor of Robert Bell building. The geometric and color appearances are very different in these two environments. The testing and training scenes are presented in Fig. \ref{fig:smil_exp}. 

\subsection{Task Description}
We have evaluated our framework on four correlated tasks that are described in Table \ref{tb:task}. Since classroom indoor environments contain fewer free space and are far more complex compared to hallway environments, we set less constraints on classroom related task, i.e. traverse classroom and to hallway. The goal of these four tasks is to simulate the robot in a multi-task decision making environment, where it is required to go to different indoor locations based on human command. 
\begin {table*}[t]
\vspace{0.1in}
\caption{Task Description}\label{tb:task}
\centering
\scalebox{0.97}{
\begin{tabular}{c|c|c|c}
\textbf{Task}&\textbf{Task Description}&\textbf{Time Limit}&\textbf{Failure Condition}\\
\hline
\hline
Traverse Hallway&  \makecell{The robot is initialized \\in hallway at a fixed position.\\ It is asked to traverse hallway \\ without collision within the time limit.} &1 min&\makecell{If the robot collides into obstacles, \\we count it as a failure. \\If it goes into classroom, \\we count it as a failure.}\\
\hline
Traverse classroom&\makecell{The robot is initialized \\in classroom at the door position. \\It is asked to traverse classroom \\ without collision within the time limit}&30 sec&\makecell{Because classroom is highly complex, \\we give the robot one more chance\\ in this task. If the robot collides into \\ obstacles, we reroute it back to \\ free space. If the robot collides\\ again, we count it as a failure. \\In addition, if it goes outside \\ of the classroom, we count it \\ as a failure. }\\
\hline
To classroom&\makecell{The robot is initialized \\in hallway at a fixed position that is 15 meters \\away from a classroom. \\ It needs to go from the \\classroom to the hallway.}&2min& \makecell{If the robot can not complete \\the task within time limit or \\ collide into obstacles, we count\\ it as a failure. In addition, if the \\robot passes two nearest classrooms \\ from its initial position, it is a failure.}\\
\hline
To hallway&\makecell{The robot is initialized \\in a classroom at the furthest corner \\away from the door. \\ It needs to go through \\the classroom to the hallway.}&1 min&\makecell{If the robot can not complete \\ the task within time limit or \\ collide into obstacles, we count it as \\ a failure. Same as traverse \\ classroom, we give the robot a \\ second chance.}\\
\hline
\end{tabular}
}
\end{table*}

\subsection{Experiment Configurations} 
A variety of baselines have been designed to evaluate the performance. We compare the $SMIL$ full architecture with five baselines:
\begin{itemize}
\item \textbf{Multi-headed network}:  this model uses the same architecture without the shared learning representation (with the addition operation excluded). 
\item \textbf{Plain network}: this is the traditional imitation learning approach, where there is one single network that maps the input to output control commands. It is not aware of the multi-task setting.
\item \textbf{$SMIL$ w/o data augmentation}: this baseline excludes data augmentation in training.
\item \textbf{$SMIL$ w/o dropout}: this baseline removes dropout at the first and second layers of sub-policies. 
\item \textbf{$SMIL$ w/o environment prediction}:  this model does not predict environment class labels. 
\end{itemize}

We use the same hyper-parameters to train each baseline network across all the tasks. We use stochastic gradient descent (SGD) optimizer with a learning rate of 0.01 and a decay factor of 10 for every 5 epochs. We train each network for 30 epochs with a batch size of 256. For the $SMIL$ framework and multi-headed network, we split training examples of each task evenly in each batch. We train the plain network on all the training data. The plain network is task-agnostic, so we do not inform it which task to perform explicitly. We set the weight decay to 0.0005. For the networks that use dropout, we set the dropout rate as 0.2 implying that with the probability of $20\%$, a neuron's activation will be set to 0.

\begin{figure*}[htp]
\centering
\includegraphics[scale=0.32]{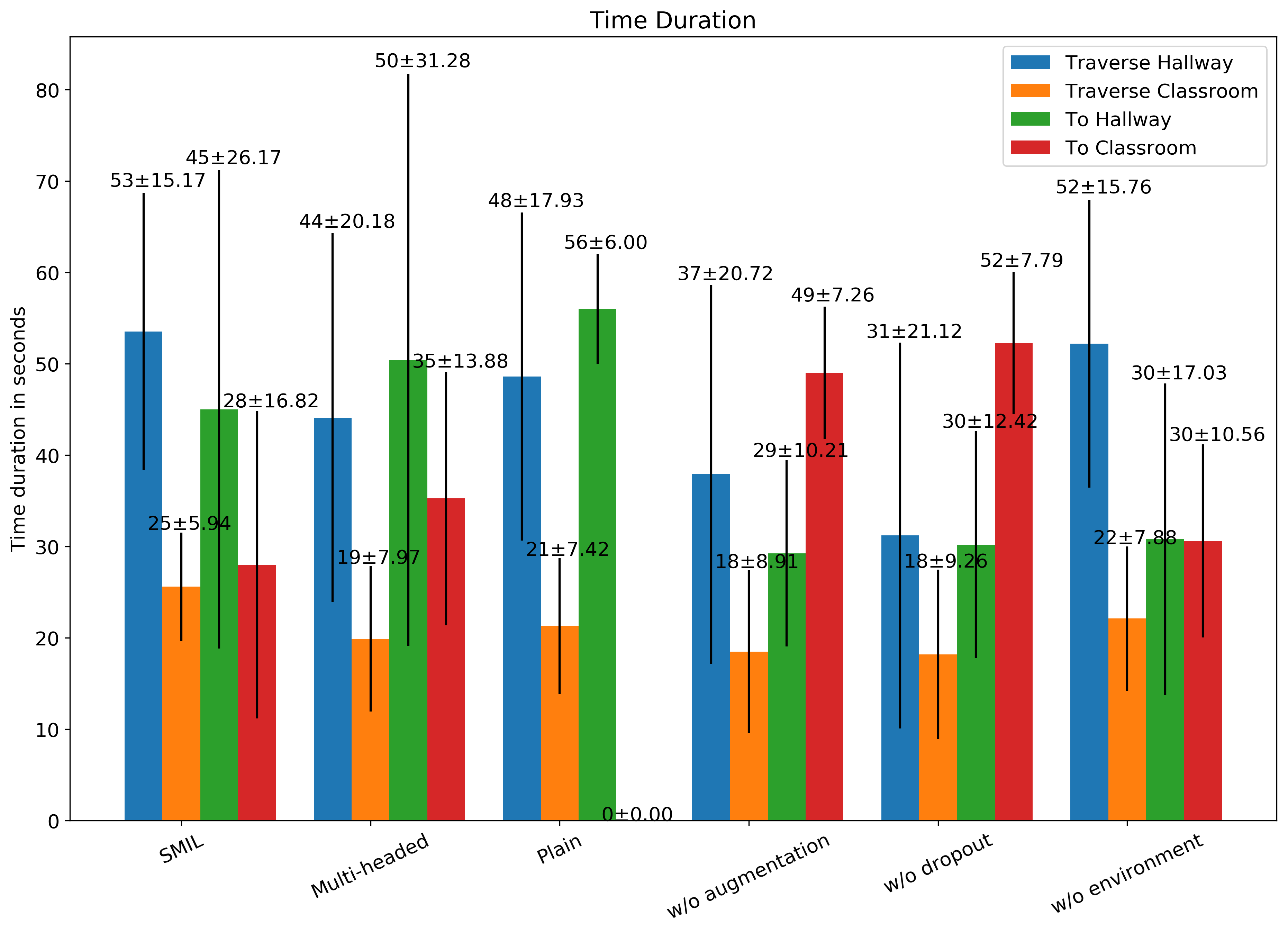}
\caption{Averaged time duration and standard deviation for each task. For traverse hallway/classroom task, The higher the averaged time duration means the the robot gets less collision. For to hallway/classroom tasks, the lower the averaged time duration means the faster the robot completes the task.}
\label{fig:time}
\end{figure*}

We have conducted 10 experiments for each task and recorded the success rate and the averaged time duration with standard deviation for each experiment as shown in Table \ref{tb:success_rate} and Fig. \ref{fig:time}.

\begin{table*}[t]
\centering
\vspace{0.3in}
\caption{Success Rate}
\label{tb:success_rate}

\scalebox{1.2}{
\begin{tabular}{c|c|c|c|c}
&Traverse Hallway&Traverse Classroom&To Classroom&To Hallway\\
\hline
\hline
$\textbf{\textit{SMIL}}$&$\textbf{80\%}$&$\textbf{60\%}$&$\textbf{70\%}$&$\textbf{60\%}$\\
Multi-headed&$50\%$&$30\%$&$50\%$&$40\%$\\
Plain&$50\%$&$30\%$&$20\%$&$0\%$\\
$SMIL$ w/o augmentation&$40\%$&$30\%$&$40\%$&$30\%$\\
$SMIL$ w/o dropout&$30\%$&$30\%$&$50\%$&$40\%$\\
$SMIL$ w/o environment&$70\%$&$50\%$&$50\%$&$50\%$\\
\hline
\end{tabular}
}
\end{table*}

\subsection{Overall Comparisons} 
As we can observe from Table \ref{tb:success_rate} and Fig. \ref{fig:time}, in terms of the success rate, it is clear that our model, $SMIL$, achieves the best performance across all the tasks. In terms of time duration, our model is able to maintain the longest averaged travels among all other models in traverse tasks. It needs slightly more time to complete to classroom task. This is because some baseline models complete all easy runs that take less time, e.g. the classrooms that are in the robot's field of view, but failed difficult runs that take longer time.

\subsection{Comparisons on Model Architectures}
The comparisons address the first three questions at the beginning of this section. We can draw three conclusions: \textbf{First}, the first floor has more obstacles than the third floor, which is used for training. Thus, it is necessary to reuse the knowledge learned from traverse classroom task while performing traverse hallway task. By comparing $SMIL$ and multi-headed network on traverse hallway task, we observe that $SMIL$ is able to reuse obstacle avoidance knowledge from traverse classroom sub-policy. \textbf{Second}, from the result, we observe that it is necessary to add the environment prediction auxiliary task to provide additional training single  to the image feature extractor, which allows the image feature extractor to learn a more robust representation. \textbf{Third}, although the plain network is trained on all the data that contains 80,000 images, it still fails tremendously especially on the tasks of {\it to classroom/hallway}. This is because the mode averaging \cite{bishop1994mixture} and inaccurate labeling cause bias to the network. The tasks of {\it to classroom/hallway} inherit a different mode from that in the {\it traverse classroom/hallway} tasks and they are more difficult because the robot needs to avoid obstacles and find the targeted place. Since the plain network is trained on all tasks, it is possible that the plain network ignores differences across each task and thus yield a poor policy. In addition, it is task-agnostic, so it can not respond to human command.

\subsection{Comparison on generalization and robustness} 
This comparison answers the fourth question: both dropout and data augmentation are necessary to train a robust model.  By removing any of them, the robot has the similar performance. In terms of the success rate, after adding both methods, the robot's performance is almost doubled. It shows that these two methods are complementary to each other. Dropout prevents the robot from aggressive turning caused by large activations from the addition operation. By augmenting the training dataset, the trained model learns to ignore geometric difference and different lighting effects.

\section{Conclusion}\label{sec:con}
In this paper, we propose a deep multi-task shared imitation learning framework, \textit{SMIL}, that can learn to work on multiple tasks with multiple sub-policies by learning the relations shared among these policies/tasks. Compared to the plain neural network, this framework allows the robot to follow human instructions. In addition, by leveraging the task relations, this framework is highly robust to new environments and produces the best results over all baselines. We have evaluated our framework in a real environment that is different from the training environment. The results show its robustness and great generalization to new environments.

\bibliographystyle{abbrv}
\bibliography{Ref.bib}
\end{document}